\documentclass[conference]{IEEEtran}
\IEEEoverridecommandlockouts

\usepackage{braket} 
\usepackage{hyperref}  
\usepackage{caption}
\usepackage{cite}
\usepackage{amsmath,amssymb,amsfonts}
\usepackage{algorithmic}
\usepackage{graphicx}

\usepackage{xcolor}
\def\BibTeX{{\rm B\kern-.05em{\sc i\kern-.025em b}\kern-.08em
    T\kern-.1667em\lower.7ex\hbox{E}\kern-.125emX}}

\title{ Fast, Accurate and  Interpretable Graph Classification with Topological Kernels}

\author{
    \IEEEauthorblockN{
        Adam Wesołowski\IEEEauthorrefmark{1}\IEEEauthorrefmark{2}, 
        Ronin Wu\IEEEauthorrefmark{2}, 
        Karim Essafi\IEEEauthorrefmark{2}
    }
    
    \IEEEauthorblockA{\IEEEauthorrefmark{1}Department of Computer Science, Royal Holloway University of London\\
    Email: \href{adam@qunasys.com}{adam@qunasys.com}}

    \IEEEauthorblockA{\IEEEauthorrefmark{2}QunaSys Europe, Copenhagen, Denmark\\
    Emails: \href{ronin@qunasys.com}{ronin@qunasys.com}, 
    \href{karim@qunasys.com}{karim@qunasys.com}}
}

\begin{document}

\maketitle

\begin{abstract}

We introduce a novel class of explicit feature maps based on topological indices that represent each graph by a compact feature vector, enabling fast and interpretable graph classification. Using radial basis function kernels on these compact vectors, we define a measure of similarity between graphs. We perform evaluation on standard molecular datasets and observe that classification accuracies based on single topological-index feature vectors underperform compared to state-of-the-art substructure-based kernels. However, we achieve significantly faster Gram matrix evaluation---up to $20\times$ faster---compared to the Weisfeiler--Lehman subtree kernel. To enhance performance, we propose two extensions: 1) concatenating multiple topological indices into an \emph{Extended Feature Vector} (EFV), and 2) \emph{Linear Combination of Topological Kernels} (LCTK) by linearly combining Radial Basis Function kernels computed on feature vectors of individual topological graph indices. These extensions deliver up to $12\%$ percent accuracy gains across all the molecular datasets. A complexity analysis highlights the potential for exponential quantum speedup for some of the vector components. Our results indicate that LCTK and EFV offer a favourable trade-off between accuracy and efficiency, making them strong candidates for practical graph learning applications.

\end{abstract}



\section{Introduction}

Graph classification is a fundamental problem in theoretical computer science and machine learning, with broad applications in chemistry~\cite{RW2,randic},  bioinformatics~\cite{bioinformatics}, and social network analysis~\cite{socialnetworks}. In this work, we focus on the classification of chemical compounds, naturally represented as undirected graphs with atoms as vertices and chemical bonds as edges.

A key challenge in graph classification lies in designing graph representations that balance expressiveness with computational efficiency. Graph Kernels—similarity functions that implicitly map graphs into high-dimensional feature spaces have proven to be effective, particularly when combined with classifiers such as Support Vector Machines (SVMs)~\cite{svm}. Among these, the Weisfeiler–Lehman (WL) and its variations have demonstrated strong empirical performance~\cite{kriege2020survey,Borgwardt_2020, nikolentzos2021graph}. However, their effectiveness often comes at the cost of computational efficiency: the WL subtree kernel, although considered to be amongst the most efficient graph kernels, produces high-dimensional feature vectors, making kernel matrix computation (e.g., Gram matrix construction) both time- and memory-demanding. In practice, this can limit their scalability on very large datasets. Recent approaches have attempted to mitigate these limitations by introducing more compact and scalable alternatives~\cite{AEGK,essafi2025benchmark,ICDM24randomwalkfingerprints}. The results in~\cite{ICDM24randomwalkfingerprints} provide a compelling argument for scalability and efficiency, however, lack on  accuracy to be comparable to the state-of-the-art methods. 


In response to these limitations, we propose a new class of topological kernels that represent each graph by a low-dimensional, explicitly computed, feature vector based on certain specific topological graph descriptors. When the underlying graph corresponds to a chemical compound these feature vectors are sometimes referred to as a {\it topological fingerprint} of a molecular graph. Some prior works~\cite{topoclass2,topoclass1} have considered topological descriptors such as average degree, graph radius or diameter, however, all approaches yielded uncompetitive accuracy results. We consider the concept of topological indices to form a graph representation. This choice is motivated by the descriptive nature of these indices, and proven ability to discriminate different types of chemical compounds~\cite{WIeneroriginal}. Examples of topological indices include the \emph{Randić index}~\cite{randic}, which is defined in terms of vertex degrees, the \emph{Wiener index}~\cite{WIeneroriginal}, based on pairwise shortest-path distances, and the \emph{Estrada index}~\cite{ESTRADA2000713}, which depends on the spectral properties of the adjacency matrix of a graph. We initially test three different methods based on three different topological indices, and report time, accuracy and $F1$ scores on some molecular datasets.
We find that feature vectors based solely on individual indices offer mediocre classification performance at significantly lower the computational cost, this is in agreement with prior work on other topological descriptors for graph classification~\cite{topoclass1,topoclass2}. The weak performance in terms of accuracy for \emph{Wiener index}, as a sole graph descriptor, has been pointed out in prior work~\cite{Borgwardt_2020}. Our findings stand in agreement with the existing belief~\cite{Borgwardt_2020} that individual topological descriptors although, usually much more computationally efficient, underperform when it comes to classification accuracy.

\begin{figure}
    \centering
    \includegraphics[width=0.9\linewidth]{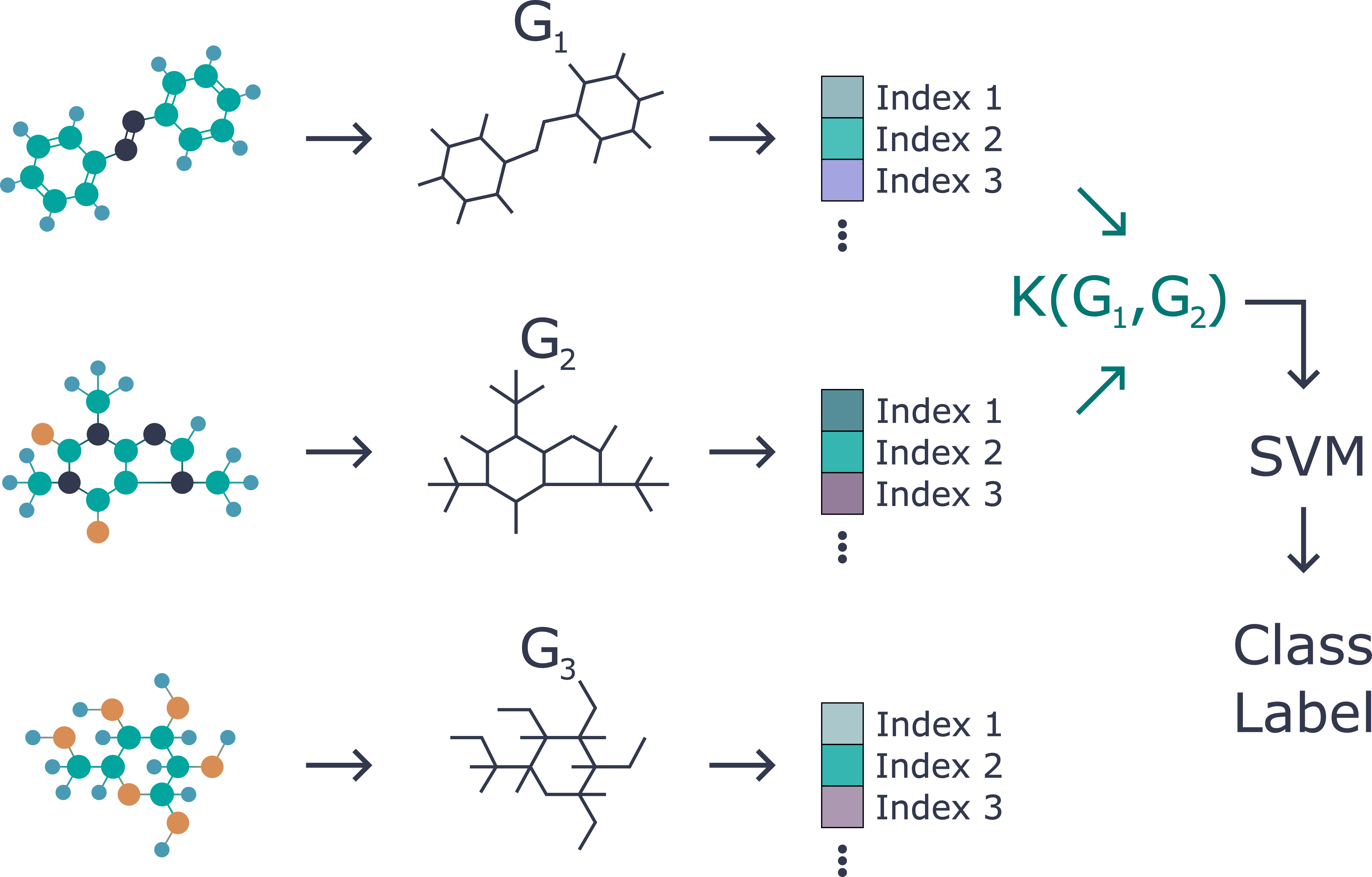}
    \caption{Visualization of molecular graph classification using topological kernels. Each graph is first abstracted into a topological fingerprint—a compact feature vector composed of graph's topological indices (ex., Wiener, Randić, Estrada). These fingerprints enable fast and interpretable evaluation of graph similarities via an RBF kernel, which introduces non-linearity and supports complex decision boundaries. The resulting kernel matrix is passed to an SVM classifier for graph classification. For illustration only one kernel evaluation between graphs $G_1$ and $G_2$ is shown.}
    \label{fig:visualisation}
\end{figure}

To improve the classification performance, we introduce two extensions. The first, called the Extended Feature Vector (EFV), concatenates multiple topological indices into a modestly higher-dimensional representation, allowing the kernel to capture richer structural patterns. The second, which we term the \textbf{LCTK}, constructs a \emph{linear combination of topological kernels} by linearly combining multiple RBF kernels, each computed on a different topological index. This formulation permits adaptive weighing of different structural descriptors and is of comparable time complexity to single index descriptors.
A visual representation of the workflow of our methods is depicted in Figure \ref{fig:visualisation}.

We demonstrate that EFV and LCTK match or exceed the accuracy of the state-of-the-art methods (including top performing variations of WL kernels) on most of the datasets, while simultaneously reducing the required computational time and memory used to perform the computation. This puts in question the need for more sophisticated approaches, when for example on $\tt{MUTAG}$ dataset the LCTK method achieves over $94.6\%$ accuracy compared to $87.6\%$ accuracy of WL subtree kernel, simultaneously LCTK computes the Gram Matrix almost $2$ times faster than the WL subtree kernel.

Beyond their practical simplicity, the EFV and LCTK methods are appealing from a computational complexity perspective. We provide a detailed analysis under both classical and quantum models, showing that topological indices are efficiently computable classically, and that several admit polynomial or exponential quantum speedups. Notably, the \emph{Estrada index} benefits from recent quantum algorithms for computing functions of eigenvalues that yield exponential performance improvements~\cite{giovannetti2025}.

We validate our approach on standard molecular graph benchmarks, including $\tt{MUTAG}$, $\tt{PROTEINS}$, $\tt{AIDS}$, $\tt{NCI1}$, and $\tt{PTC-MR}$. The results show up to a $15\times$ speedup in Gram matrix computation over the WL subtree kernel~\cite{Shervashidze2011WeisfeilerLehmanGK}. While being significantly more computationally efficient, the EFV and LCTK extensions achieve accuracy gains of up to $12\%$ over single-index descriptors, surpassing both on accuracy and time the WL subtree kernel on 3 out of 6 benchmarked datasets. These findings suggest that low dimensional topological index-based embeddings provide a compelling balance of accuracy, interpretability, and efficiency, and may offer a viable alternative to more resource-intensive graph learning techniques, even on classical hardware.

\section{Preliminaries}

\subsection{Graph Kernels vs.\ Kernelized Topological Descriptors}

In this section, we contrast two paradigms for measuring similarity between graphs in supervised learning: \emph{graph kernels}, which compare substructures of graphs directly, and \emph{kernelized topological descriptors}, which first extract summary vectors and then apply a generic kernel (e.g.\ Gaussian, polynomial or linear) over those vectors.

\subsection{Graph Kernels}
Graph kernels define a positive–definite function 
\[
k\colon \mathcal{G}\times\mathcal{G}\to\mathbb{R}
\]
directly on the space \(\mathcal{G}\) of graphs by implicitly embedding each graph \(G\) into a (potentially infinite–dimensional) reproducing kernel Hilbert space (RKHS) \(\mathcal{H}\) via a map \(\phi:G\mapsto\phi(G)\in\mathcal{H}\), such that
\[
k(G,G') = \bigl\langle \phi(G),\,\phi(G')\bigr\rangle_{\mathcal{H}}.
\]
These kernels exploit the \emph{kernel trick} to avoid explicit feature construction and can capture rich structural information. These approaches, however,  very often require prohibitive computational time.

\subsection{Kernelized Topological Descriptors}
An alternative is to compute explicitly a small number \(d\) of \emph{topological descriptors} for each graph and concatenate them into a single feature vector:
$$
    \psi(G) = \bigl[\,\mathrm{TD_1}(G),\,\mathrm{TD_2}(G),\,\mathrm{TD_3}(G),\,\dots\bigr] \in \mathbb{R}^d,
$$
$$
k_{\mathrm{topo}}(G,G') \;=\; \exp\!\bigl(-\gamma\,\|\psi(G)-\psi(G')\|^2\bigr)$$

This two‐step approach has advantages:
\begin{itemize}
  \item \emph{Simplicity and interpretability}: descriptors are intuitive, low–dimensional summaries.
  \item \emph{Modularity}: any feature‐space kernel can be swapped in (RBF, polynomial, etc.).
  \item \emph{Efficiency}: feature‐vector construction is often an efficient processs; RBF evaluation is \(\mathcal{O}(1)\), for vectors of constant dimension.
\end{itemize}
However, this approach may lose fine‐grained structural distinctions that sophisticated graph kernels capture, and performance hinges on descriptor choice and kernel hyperparameters.
Kernelized topological descriptors suit scenarios demanding scalability, interpretability, or when prior knowledge suggests a small set of discriminative graph invariants. For a more detailed discussion of implicit versus explicit feature maps we refer the reader to\cite{implicitvsexplicit}.

\subsection{The Radial Basis Function (RBF) Kernel}

A widely used kernel is the Radial Basis Function (RBF) kernel, also known as the Gaussian kernel, defined as:

\begin{equation}
k(\mathbf{x}, \mathbf{x}') = \exp\left( -\gamma\|\mathbf{x} - \mathbf{x}'\|^2 \right),
\end{equation}

where $\gamma > 0$ is the kernel width parameter that controls the shape of the decision boundary.

Empirical results show that, the RBF kernel can significantly enhance classification performance compared to a linear kernel\cite{RBFvslinear,rbflinear2}. 
\subsection{Support Vector Machines (SVMs)}

SVMs are large-margin classifiers that find the hyperplane maximizing the margin between classes. In the separable case, given labeled training data $\{ (\mathbf{x}_i, y_i) \}_{i=1}^n$, where $y_i \in \{-1, +1\}$, the primal optimization problem is:

\begin{equation}
\begin{aligned}
\min_{\mathbf{w}, b} \quad & \frac{1}{2} \|\mathbf{w}\|^2 \\
\text{s.t.} \quad & y_i (\mathbf{w}^\top \mathbf{x}_i + b) \geq 1, \quad \forall i.
\end{aligned}
\end{equation}

For non-separable data, slack variables $\xi_i \geq 0$ are introduced, leading to the soft-margin formulation:

\begin{equation}
\begin{aligned}
\min_{\mathbf{w}, b, \boldsymbol{\xi}} \quad & \frac{1}{2} \|\mathbf{w}\|^2 + C \sum_{i=1}^n \xi_i \\
\text{s.t.} \quad & y_i (\mathbf{w}^\top \mathbf{x}_i + b) \geq 1 - \xi_i, \quad \forall i, \\
& \xi_i \geq 0, \quad \forall i,
\end{aligned}
\end{equation}

where $C > 0$ controls the trade-off between margin maximization and training error.

Using the representer theorem, the dual problem becomes:

\begin{equation}
\begin{aligned}
\max_{\boldsymbol{\alpha}} \quad & \sum_{i=1}^n \alpha_i - \frac{1}{2} \sum_{i,j=1}^n \alpha_i \alpha_j y_i y_j k(\mathbf{x}_i, \mathbf{x}_j) \\
\text{s.t.} \quad & 0 \leq \alpha_i \leq C, \quad \forall i, \\
& \sum_{i=1}^n \alpha_i y_i = 0.
\end{aligned}
\end{equation}

The decision function for a new input $\mathbf{x}$ is:

\begin{equation}
f(\mathbf{x}) = \text{sign} \left( \sum_{i=1}^n \alpha_i y_i k(\mathbf{x}_i, \mathbf{x}) + b \right),
\end{equation}

where the support vectors correspond to training points with $\alpha_i > 0$.

\subsection{Topological Indices}

\subsubsection{Wiener Index}
The \emph{Wiener index}~\cite{WIeneroriginal} of a molecular graph $G=(V,E)$ with vertex set $V$ and edge set $E$ is given by
\[
W(G) \;=\; \sum_{\{u,v\}\subset V} d_G(u,v),
\]
where $d_G(u,v)$ denotes the shortest-path distance between vertices $u$ and $v$ in $G$. Historically one of the earliest distance-based indices, $W(G)$ exhibits strong linear correlations with boiling points~\cite{WIeneroriginal} and other physico-chemical properties of alkanes and related compounds.\cite{rouvray1976dependence}

\subsubsection{Estrada Index}
The \emph{Estrada index}~\cite{ESTRADA2000713} leverages the spectrum of the adjacency matrix $A$ of $G$. If the eigenvalues of $A$ are $\lambda_1,\dots,\lambda_n$, then
\[
E(G) \;=\; \sum_{j=1}^n e^{\lambda_j}
\;=\;\mathrm{tr}\bigl(e^A\bigr).
\]
This spectral measure captures global connectivity and subgraph centralities simultaneously. In QSPR studies, $E(G)$ has been shown to enhance models for heats of formation and entropy when used alongside other eigenvalue-based descriptors.

\subsubsection{Randić Index}
The \emph{Randić index}~\cite{randic} quantifies molecular branching via the degree sequence of $G$. It is defined as
\[
R(G) \;=\; \sum_{uv \in E} \frac{1}{\sqrt{\deg(u)\,\deg(v)}},
\]
where $\deg(v)$ is the degree of vertex $v$. The \emph{Randić index} is correlated with the boiling point for alkanes~\cite{randic} and some X-ray absorption parameters~\cite{Khatri_2016}.

\section{Similarity measure with Topological Indices}  

Traditional graph kernels, such as the random walk kernels~\cite{kriege2020survey} or WL subtree kernel~\cite{wlsubtree}, often generate high-dimensional feature representations, leading to increased computational complexity of computing the Gram matrix.

An alternative approach is to construct a \textbf{topological-index kernel}, where a graph is represented as a one-dimensional feature vector whose sole entry is a topological index of the graph. Formally, given a graph $ G $, we define the feature map  

\begin{equation}
\phi(G) = (T(G)),
\end{equation}  

where $ T(G) $ is a chosen topological index, such as the \emph{Wiener index}, \emph{Estrada index}, or \emph{Randić index}. The resulting \emph{topological kernel} is then defined as  

\begin{equation}
K(G_1, G_2) = e^{ -\gamma(||T(G_1) -T(G_2)||^2)}.
\end{equation}  

By leveraging a single numerical descriptor, this approach offers a compact and interpretable representation of graph structure, substantially reducing the dimensionality of the feature space. In contrast to traditional graph kernels, which typically rely on comparisons of substructures, a topological index encapsulates global graph properties into a single value, thereby facilitating a more efficient classification process. A key advantage of using topological indices lies in their simplicity and invariance under graph isomorphism (for unlabeled or unattributed graphs): isomorphic graphs necessarily share identical values of any topological index. However, topological indices are not complete invariants; distinct non-isomorphic graphs may, in some cases, yield identical or nearly identical index values.

Despite certain limitations, topological indices offer significant utility, particularly in scenarios where computational efficiency is crucial. The representation via topological indices yields a fixed-dimensional vector, independently of the input size. This property facilitates rapid (i.e. constant time) arithmetic operations during classification, as manipulating constant-sized vectors is substantially more efficient than operating on vectors whose dimensions scale polynomially with the size of the graph, i.e. $poly(n)$.
Topological indices thus provide a low-dimensional, easily computable, and interpretable representation of graphs. Their graph-invariant nature ensures that they retain meaningful discriminative power while avoiding the computational burden associated with high-dimensional feature spaces. This approach is particularly advantageous in settings where graphs exhibit pronounced structural patterns that are well-captured by topological measures. Moreover, by drastically reducing the cost of kernel evaluations, it renders large-scale graph classification tasks considerably more tractable.

\subsection{Classical complexity of computing topological index kernels.}
To compute \emph{Wiener index} one needs to find all-pairs-shortest-paths which has an established time complexity of $O(mn)$ where $m$ is the number of edges, and $n$ is the number of nodes in the graph. One may simply use Dijkstra algorithm~\cite{dijkstra1959note} from every vertex in a graph and sum up the lengths of all $n(n-1)/2$ shortest paths between all pairs of vertices in $O(mn+n^2)$ time. 
\emph{Randić index} is computing a function of degrees of pairs of vertices and then summing up the results. This requires at most $O(n\cdot\Delta_{max})=O(m)$ operations, where $\Delta_{max}$ is the maximal vertex degree in a graph. 
\emph{Estrada index} comes down to computing the trace of the exponentiated adjacency matrix of the corresponding graph. Computing trace classically requires $O(n^3)$ time. One can also perform full eigendecomposition and sum up the exponentiated eigenvalues which also takes $O(n^3)$ time.

\subsection{Quantum complexity of computing topological kernels.}
The best known quantum algorithm to compute all-pairs-shortest-path (SSSP) is based on the Quantum Dijkstra algorithm~\cite{durr2006quantum} which requires $\tilde{O}(\sqrt{nm})$, thus the total time to compute SSSP from every vertex  and sum up lengths of all-pairs-shortest-paths in a graph in a quantum setting is $\tilde{O}(n\sqrt{nm}+n^2)$. This constitutes a polynomial quantum speedup. On specific graph instances one may also use the quantum algorithm for single-pair-shortest path\cite{wesolowski2024advances} to approximate the \emph{Wiener index}. The \emph{Randić index} \( R(G) = \sum_{u,v \in E(G)} \frac{1}{\sqrt{d(u)d(v)}} \) requires access to the degrees of adjacent vertices across all edges in the graph. Consequently, any algorithm—classical or quantum—must process each edge to compute the index accurately, imposing a necessity of $ O(m) $ algorithmic steps, where $ m $ is the number of edges in the graph. Quantum algorithms do not offer a speedup for this computation, as the process fundamentally involves aggregating local degree information across the entire edge set. Therefore, the computation of the \emph{Randić index} remains a linear-time operation in both classical and quantum paradigms.
Interestingly, it has been proven by~\cite{giovannetti2025quantum, luongo2020quantum} that computing the determinant of a sparse matrix to accuracy $\epsilon$ admits an exponential quantum speedup, i.e. can be done in $O(\log{n}/\epsilon^3)$, compared to $O(n^3)$ classical runtime. Thus computing \emph{Estrada index} admits exponential quantum speedup and thereby is of the highest interest in terms of accuracy results.

\renewcommand{\arraystretch}{1.5} 
\begin{table*}[ht]
\centering
\begin{tabular}{c|ccccc}

 & Wiener & Estrada & Randić & EFV & LCTK \\
\hline

 Classical& $O(mn)$ & $O(n^3)$ & $O(m)$ & $O(n^3)$  &  $O(n^3)$ \\

 Quantum & $\tilde{O}(n^{1.5}\sqrt{m})$ &$O(\log{n}/\epsilon^3)$  & $O(m)$  &  $\tilde{O}(n^{1.5}\sqrt{m})$ &  $\tilde{O}(n^{1.5}\sqrt{m})$ \\

\hline
\end{tabular}
\caption{Time complexity of different methods for computing feature vectors introduced in this work.}
\label{tab:empty-table}
\end{table*}

\section{Extension I: Extended Feature Vector (EFV)}

Consider the alternative strategy of concatenating topological indices into a feature vector $\mathbf{f}(G) = [f_1(G),\ldots,f_m(G)]$ with a single RBF kernel:

\begin{equation}
K_{\text{EFV}}(G,G') = \exp\left(-\gamma \|\mathbf{f}(G) - \mathbf{f}(G')\|^2\right)
\end{equation}

\subsection{EFV vs single index kernels }
The following points provide a brief comparative analysis of \emph{Extended Feature Vector} approach and the single index kernels:
\begin{itemize}
    \item \textbf{Expressivity:} Extended feature vectors offer significantly richer representations, as they combine diverse structural summaries. This can improve classification performance, particularly when the dataset contains subtle structural differences not captured by a single index.
    
    \item \textbf{Robustness:} Using multiple graph indicies reduces the risk of over-reliance on a potentially non-informative topological index.
    
    \item \textbf{Simplicity and Speed:} The single-index kernel has minimal overhead and is particularly well-suited for massive datasets or scenarios where quick computation is critical.

    \item \textbf{Scalability:} When $n$ is large, storing and manipulating the full kernel matrix becomes a bottleneck. Both single-index kernels and EFV provide a lightweight technique with excellent scalability properties.
\end{itemize}

\section{Extension II: Linear Combination of Topological Kernels (LCTK) }
\label{sec:kernel_combinations}

The classification of graphs via topological indices can be enhanced through a weighted combination of individual topological kernels. Given a collection of graphs $\mathcal{G} = \{G_1,\ldots,G_n\}$, we consider $m$ topological indices $f_1,\ldots,f_m$ where each $f_j:\mathcal{G} \rightarrow \mathbb{R}$ computes a specific graph invariant.

\subsection{Kernel Formulation}
For each topological index $f_j$, we define an associated kernel function $k_j$ that measures similarity between graphs. The linear combination approach constructs a composite kernel as a convex combination:

\begin{align}
\begin{split}
    & K_{\text{LCTK}}(G,G') = \sum_{j=1}^m w_j k_j(G,G'), \\
& \quad \text{with } w_j \geq 0 \text{ and } \sum_{j=1}^m w_j = 1
\end{split}
\end{align}

\subsection{Comparative Analysis of LCTK}
The LCTK approach offers several theoretical and practical benefits compared to using individual topological indices or EFV approach.
The kernel combination approach differs in three key aspects:

\begin{enumerate}
\item \textbf{Scale Sensitivity}: The combined kernel applies separate bandwidth parameters $\gamma_j$ for each index, automatically adapting to their different scales. The concatenated (EFV) approach requires careful normalization as all dimensions share one $\gamma$.

\item \textbf{Interpretability}: The weights $w_j$ in $K_{\text{LCTK}}$ directly indicate each index's contribution. The EFV obscures individual index importance.

\item \textbf{Adaptive Weighting}: The coefficients $w_j$ can be optimized to emphasize the most discriminative indices for a given classification task. This provides flexibility that single-index and EFV approaches lack.

\item \textbf{Enhanced Expressivity}: Different topological indices capture distinct aspects of graph structure. The \emph{Wiener index} reflects global connectivity patterns, while the \emph{Estrada index} encodes spectral information and the \emph{Randić index} characterizes local vertex interactions. The linear combination integrates these complementary perspectives.

\end{enumerate}

Importantly, LCTK kernel as a linear combination of RBF kernels retains positive-definiteness and thus constitutes a valid kernel approach, compatible with the SVM workflow.

\section{Methodology}
  For the SVM classification datasets were split into $80\%$ training $20\%$ test subsets. Chemistry datasets were taken from TU Dortmund repository~\cite{KKMMN2016}. Graphs were generated using networkx Python library~\cite{networkx}. The computation was performed on MacBook Air with 1,1 GHz Dual-Core Intel Core i3 processor and 8 GB of Random Access Memory.

  We benchmark our methods on the following datasets:
  \begin{itemize}
  \item \textbf{AIDS}: A dataset comprising $2000$ molecular graphs derived from the Antiviral Screen Database. Each graph represents a chemical compound, with nodes as atoms and edges as bonds. The classification task involves predicting the anti-HIV activity of these compounds.~\cite{aids}
  
  \item \textbf{NCI1}: Contains $4110$ chemical compound graphs from the National Cancer Institute's screening tests. Each graph represents a molecule, with nodes as atoms and edges as chemical bonds. The objective is to classify compounds based on their activity against non-small cell lung cancer and ovarian cancer cell lines.~\cite{ncl1}
  
  \item \textbf{PTC-MR}: Comprises $344$ chemical compound graphs labeled according to their carcinogenicity on male rats. Nodes represent atoms, and edges denote chemical bonds. The dataset is used to predict the carcinogenic potential of compounds.~\cite{ptc-mr}
  
  \item \textbf{MUTAG}: Consists of $188$ nitroaromatic compounds represented as graphs, where nodes are atoms and edges are chemical bonds. The classification task is to predict the mutagenic effect of these compounds on \emph{Salmonella typhimurium}.~\cite{Debnath1991StructureactivityRO}
  
  \item \textbf{PROTEINS}: Contains $1113$ protein graphs, with nodes representing amino acids and edges indicating spatial proximity. The task is to classify proteins as enzymes or non-enzymes based on their structure.~\cite{RW2}
  
  \item \textbf{DD}: A dataset of $1178$ protein structure graphs, where nodes represent amino acids and edges denote spatial or sequential proximity. The classification goal is to distinguish enzymes from non-enzymes.~\cite{DOBSON2003771}
\end{itemize}

\subsection{Parameter Search}
For the \emph{Linear Combination of Topological Kernels} approach (LCTK) 
the experiments explored:
\begin{itemize}
    \item \textbf{Weights}: 7 combinations emphasizing different indices:
    \begin{itemize}
        \item Equal weights: $[1/3, 1/3, 1/3]$;
        \item Wiener-emphasized: $[0.5, 0.3, 0.2]$;
        \item Estrada-emphasized: $[0.2, 0.5, 0.3]$;
        \item Randić-emphasized: $[0.3, 0.2, 0.5]$;
        \item Strong single-index weights:\\ $[0.8, 0.1, 0.1]$, $[0.1, 0.8, 0.1]$, $[0.1, 0.1, 0.8]$.
    \end{itemize}
    
    \item \textbf{SVM Parameters}:
    \begin{itemize}
        \item Regularization: $C \in \{10^{-4}, 10^{-3}, \dots, 10^4\}$
        \item Kernel bandwidth: $\gamma \in \{10^{-4},10^{-3}, \dots, 10^{4}\}$
    \end{itemize}
\end{itemize}

\subsubsection{Validation}
Performance was evaluated via \textbf{stratified 10-fold cross-validation}, reporting accuracy and F1 scores with standard deviations. Each configuration was tested across all weight combinations and parameter values, amounting to the total search space size of 567 possible combinations.

\section{Results \& Analysis}

\label{sec:results}
Table II. lists accuracy results for individual topological kernels (\emph{Wiener index}, \emph{Estrada index}, \emph{Randić index}) and the two proposed extensions, namely \emph{Extended Feature Vector} (EFV) and \emph{Linear Combination of Topological Kernels} (LCTK).

Table III contains F1 scores for all of the methods introduced in this work. Table IV outlines the times required to compute the Gram matrix on different datasets using methods introduced in this work as well as WL subtree kernel and Shortest Path Kernel.
Our experiments reveal several insights about the topological kernels:
\begin{itemize}
    \item  LCTK approach is on average $2.2$ times faster compared to the  WL subtree kernel (for Gram matrix computation, averaged over all datasets excluding DD) (see Table IV);
    \item EFV is faster than LCTK on all datasets, has comparable F1 scores but tends to fall behind on accuracy;
    \item LCTK or EFV provide best F1 score on 6 out of 6 datasets, outperforming WL subtree and SP kernels.
\end{itemize}

\subsection{Performance Characteristics}
The proposed kernels exhibit distinct performance profiles across datasets. Below we provide a comparative analysis of performance of our methods against the WL subtree kernel. In particular, we highlight the relative performance of our best performing EFV and LCTK methods vs the WL subtree kernel.
On \texttt{MUTAG} the LCTK approach achieves state-of-the-art results ($94.6\%$ accuracy), demonstrating that a small change to a kernel may substantially affect the classification results without substantially impeding the runtime. \\
For \texttt{PROTEINS}, all variants maintain stable performance between $71\%-75\%$ accuracy while being $3$ to $8$ $\times$ faster than WL subtree kernel, showing scalability advantages without accuracy trade-offs.\\
For \texttt{PTC-MR} single topological index kernels provide lower (by $\pm$ $5$-$8$$\%$) accuracy comapred to WL subtree kernel. Nevertheless, both EFV and especially LCTK method outperfrom WL subtree kernel by $5\%$ and $6\%$, respectively. At the same time EFV computes the Gram matrix $7.22$ times faster than WL subtree approach, and LCTK $4.28$ times faster than WL subtree approach. On the \texttt{DD} dataset, consisting of large molecular graphs of proteins, the \emph{Estrada index} appears computationally challenging, impacting the time of the EFV and LCTK methods substantially. However, surprisingly the individual \emph{Randić index} kernel achieves over $80\%$ F1 score almost matching the best result (achieved by LCTK) for the \texttt{DD} dataset, and is more than $19$ times faster than LCTK approach (due to high time for \emph{Estrada index}). We note however, that on quantum computers this problem should be alleviated since quantum time complexity computing of computing \emph{Estrada index} is merely $O(\log{n})$.

\subsection{Computational Efficiency on Large Random Graphs}
\begin{figure}[ht]
    \centering
    \includegraphics[width=0.8\linewidth]{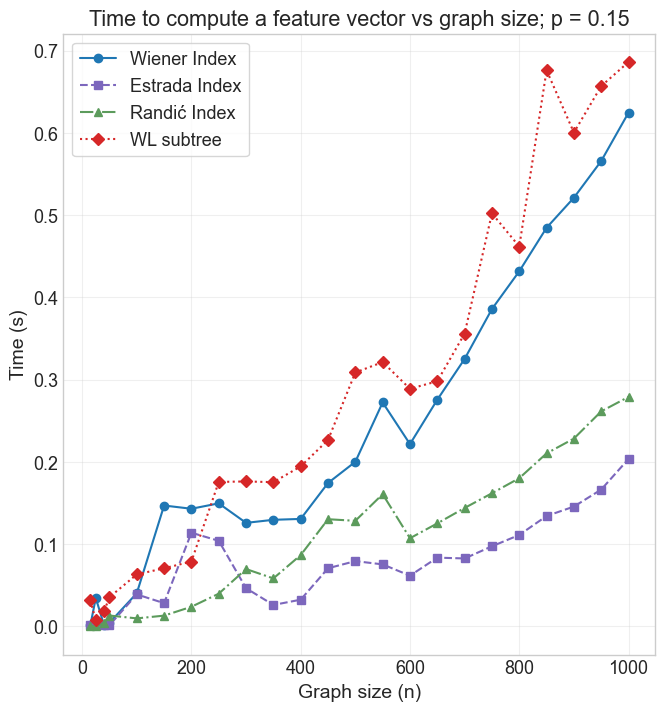}
    \caption{Time required to compute a feature vector for  each method on Erdős–Rényi random graphs with $p=0.15$, corresponding to sparse random graphs. The $h$ parameter in WL subtree kernel is set to $h=7$.}
    \label{fig:sparse}
\end{figure}

\begin{figure}[ht]
    \centering
    \includegraphics[width=0.8\linewidth]{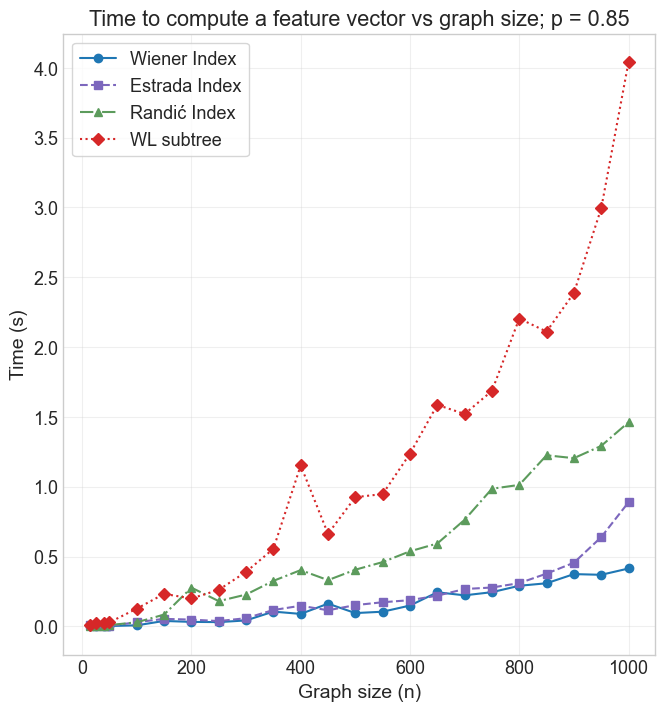}
    \caption{Time required to compute a feature vector for  each method on Erdős–Rényi random graphs with $p=0.85$, corresponding to dense random graphs. The $h$ parameter in WL subtree kernel is set to $h=7$.}

    \label{fig:dense}
    
\end{figure}

The kernels demonstrate remarkable scalability on both dense and sparse random graphs. Figures \ref{fig:sparse} and \ref{fig:dense} depict time required to compute feature vectors on sparse and dense Erdős–Rényi random graphs\cite{erdos_renyi}, respectively. From Figures \ref{fig:sparse} and \ref{fig:dense} one may infer that on sparse graphs (often encountered in chemoinformatics) \emph{Randić index} and \emph{Estrada index} (and thereby corresponding EFV or LCTK) approaches have demonstrably better scalability than WL subtree kernel. Whereas, on dense graphs (often encountered in social network graphs) \emph{Wiener index} and \emph{Estrada index} appear to have favorable scalability, with \emph{Randić index} performing similarly to the scaling of WL subtree approach.

\subsection{Theoretical Implications}
\begin{itemize}
    \item The strong performance of two out of three individual indices on \texttt{AIDS} ($\approx99\%$ accuracy) \texttt{MUTAG} ($\approx82.6\%$ accuracy) and \texttt{DD} ($\approx 75\%$ accuracy) challenges the necessity of complex feature maps for certain graph classes.
    \item The consistent improvement of LCTK (up to 12\%) over individual indices across all datasets indicates weighted combinations of single-index kernels may result in sufficiently expressive kernels, countering a common belief that hand-picked topological descriptors lack expressivity.
\end{itemize}

\begin{table*}[ht]
\centering

\begin{tabular}{|c|c|c|c|c|c|c|}
\hline
 & $\tt{AIDS}$ & $\tt{NCL1}$ & $\tt{PTC-MR}$ & $\tt{MUTAG}$ & $\tt{PROTEINS}$ & $\tt{DD}$ \\
\hline
Wiener Index Kernel (this work) & $95.8\pm 0.9$ & $65.7\pm0.8$& $58.4\pm 9.9$ & $74.3\pm 1.3$ & $71.1\pm 3.4$&$68.7 \pm 3.9$ \\
Estrada Index Kernel (this work) & $99.5\pm 0.3$ & $67.1\pm2.0$ &$57.0\pm 9.2$  & $83.6\pm 5.8$ & $71.0\pm 1.5$&$74.4 \pm 3.1$ \\
Randić Index Kernel (this work) & $98.8\pm 0.3$ & $66.2\pm0.2$ & $61.3\pm7.3$ & $81.6\pm 3.7$ & $70.0\pm 4.4$ & $75.7 \pm 1.6$\\
EFV (this work)&$99.7\pm0.05$&$66.5\pm1.2$&$69.6\pm7.7$&$92.1\pm7.5$&$75.3\pm 3.7$& $76.3 \pm 2.2$\\
\textbf{LCTK} (this work) &$99.6\pm0.1$&$65.0\pm1.9$&$71.0\pm7.5$&$94.6\pm5.1$&$75.7\pm3.6$& $76.4 \pm 1.9$\\

RW kernel & $79.7\pm 2.3$ & $\tt{timeout}$ & $54.4\pm 9.8$ & $81.4\pm 8.9$ & $69.5\pm5.1$&out of mem \\
SP kernel & $99.3\pm 0.4$ & $72.5\pm2.0$ & $60.2\pm9.4$ & $82.4\pm5.5$ & $74.9\pm3.2$ & $77.9\pm 4.5$\\
WL Subtree kernel & $98.3\pm 0.8$ & $85.2\pm 2.2$ & $64.9\pm 6.4$ & $86.7\pm 7.3$ & $76.2\pm 3.5$ & $78.7\pm 2.3$\\
DGCNN &$99.1\pm1.4$ &$76.4\pm 1.7$ &$59.5\pm6.9$ &$84.0\pm7.1$ &$73.2\pm3.2$&$76.6\pm4.3$  \\
GIN & $98.8 \pm  0.6$&$80.0\pm 1.4$ &$59.1 \pm 7.0$ &$84.7 \pm 6.7$ &$72.8 \pm 3.6$ &$75.3\pm2.9$\\
DiffPool & $99.2 \pm 0.3$ &$76.9 \pm 1.9$ &$61.1 \pm5.6$ & $79.8 \pm 6.7$&$72.5 \pm 3.5$ &$75.0\pm3.5$ \\
\hline
\end{tabular}

\caption{Comparison of the accuracy results (in \%) of the the best performing method introduced in this work: \emph{Linear Combination of Topological Kernels} (LCTK), Extended Feature Vector (EFV) kernel, and  individual index kernels with: Random Walk (RW) kernel, Shortest Path (SP) kernel,  Weisfeiler-Lehman kernel (WL subtree), Deep Graph Convolutional Neural Network (DGCNN)~\cite{dgcnn}, Graph Isomorphism Network (GIN)~\cite{gin}, and Differentiable Graph Pooling (DiffPool)~\cite{ying2018diffpool}. The best performing method is highlighted. All the results for methods not introduced in this work are taken from~\cite{nikolentzos2021graph}. For comparison of accuracy score to other kernels (including other variants of WL kernel) we refer the reader to~\cite{nikolentzos2021graph}. }
\end{table*}
\begin{table*}[ht]
\centering

\begin{tabular}{|c|c|c|c|c|c|c|}
\hline
 & $\tt{AIDS}$ & $\tt{NCL1}$ & $\tt{PTC-MR}$ & $\tt{MUTAG}$ & $\tt{PROTEINS}$ &$\tt{DD}$ \\
\hline
Wiener Index Kernel (this work) &$97.3$&$55.1 \pm 0.1$&$35.2 \pm 17.4$&$81.9 \pm 6.2$&$70.4\pm3.9$& $72.7 \pm 0.1$ \\
Estrada Index Kernel (this work)&$99.7 \pm 0.3$&$59.05 \pm 1.7$ &$42.0 \pm 16.38$&$86.8 \pm 5.7$&$79.1 \pm 2.6$ & $79.7 \pm 0.1$\\
Randić Index Kernel (this work)& $98.4 \pm 0.5$ &$58.8 \pm 2.1$&$43.9 \pm 15.3$&$87.3 \pm 6.5$&$ 77.5 \pm 3.7$& $80.6 \pm 0.1$ \\
EFV (this work)&$99.8 \pm 0.3$&$66.7 \pm 0.1$&$48.2 \pm 1.4$&$89.0 \pm 4.7$ &$79.1 \pm 2.8$&$80.4 \pm 1.7$\\
\textbf{LCTK} (this work)&$99.8 \pm 0.4$&$66.7 \pm 0.1$ &$50.4 \pm 1.4$&$88.9 \pm 5.8$&$79.3 \pm 2.6$&$80.7 \pm 1.3$ \\

SP &$91.94 \pm 0.9$&$43.2 \pm 4.7$&$43.8 \pm 0.9$&$79.9 \pm 1.7$&$74.8 \pm2.1$& $75.1 \pm 0.1$\\
WL Subtree & 96.1 & $54.9 \pm 0.1$&$41.5 \pm 10.9$& $78.0 \pm 11.1$&$75.6 \pm 3.5$& $74.1 \pm 0.1$\\

\hline
\end{tabular}

\caption{Comparison of the F1-score results of the individual index kernels, Extended Feature Vector (EFV) kernel, \emph{Linear Combination of Topological Kernels} (LCTK), Shortest Path (SP) kernel and the Weisfeiler-Lehman kernel (WL subtree). The best performing method is highlighted.}
\label{table:F1-Score}
\end{table*}

\begin{table*}[ht]
\centering
\scalebox{1}{
\begin{tabular}{|c|c|c|c|c|c|c|}
\hline
 & $\tt{AIDS}$ & $\tt{NCL1}$ & $\tt{PTC-MR}$ & $\tt{MUTAG}$ & $\tt{PROTEINS}$&$\tt{DD}$  \\
\hline
Wiener Index Kernel (this work) &0.24  ±  0.08   &1.31 ± 0.32&0.03 ± 0.01  &0.02 ± 0.01 &0.57 ±0.1&$4.58\pm0.26$\\
Estrada Index Kernel (this work) & 0.39 ± 0.15 &1.16 ± 0.08&0.05 ± 0.1  &0.03 ± 0.01  & 0.53 ±0.08&$12.73\pm0.46$\\
\textbf{Randić Index Kernel} (this work) & 0.14 ± 0.05 &0.59 ± 0.05& 0.02 ± 0.01 & 0.01 ± 0.005 & 0.18± 0.02&$ 0.93 \pm0.89 $\\

EFV&0.66± 0.23&2.66 ± 0.39 &0.01 ± 0.01 &0.05 ± 0.01& 1.24 0.18&$18.19\pm0.73 $\\
LCTK&0.81 ± 0.28&3.22 ±0.49&   0.10 ± 0.01 & 0.06 ± 0.01& 1.32 ±0.19&$18.25\pm0.74 $\\

WL (Subtree) Kernel  &2.13 ± 0.63   &11.5 ± 2.77 &0.23 ± 0.13 & 0.09  ± 0.04&1.48± 0.15&$5.12\pm 0.18$\\

SP Kernel &  0.50 ± 0.09  &2.80 ± 0.26 & 0.1 ± 0.01 & 0.05 ± 0.01 & 1.18 ± 0.1&$15.26\pm  0.88 $\\

\hline
\end{tabular}
}
\caption{Comparison of the times (in seconds) used for the kernel matrix computation on (classical).  Averaged over 20 repetitions. This table serves as an illustration of the efficiency of using constant sized feature vectors. The best performing method is highlighted.}
\end{table*}


\section{Discussion}

This study introduces a new framework for graph classification based on topological index kernels. 

The foundational approach-using individual indices as standalone descriptors-demonstrates that topological kernels offer significant computational advantages, both theoretically and empirically. However, as demonstrated in our benchmarks, single-index kernels may underperform in classification accuracy compared to methods like the Weisfeiler-Lehman (WL) subtree kernel. To address this, we introduced two extensions: the \emph{Extended Feature Vector} (EFV) and \emph{Linear Combination of Topological Kernels} (LCTK). These methods increase expressivity by combining multiple indices, either as concatenated feature vectors or as convex topological kernel mixtures.

Empirical evaluations indicate that both EFV and LCTK consistently improve classification accuracy across all benchmarked datasets, with improvements reaching up to 12\%. Notably, these enhancements do not compromise computational efficiency. On most datasets, EFV and LCTK not only outperforms the WL kernel in classification accuracy and F1 score but also in runtime.

Beyond performance, one of the most appealing features of our approach is its interpretability. Each dimension of a topological fingerprint corresponds to a graph invariant, such as the \emph{Wiener}, \emph{Estrada}, or \emph{Randić index}. This transparency allows a direct link of classification outcomes to known structural properties of the graphs- an advantage not afforded by many state-of-the-art graph classification techniques that function as black-box methods. This is particularly valuable in domains where understanding the rationale behind predictions is crucial, such as in biomedical applications or critical infrastructure systems.

Furthermore, our complexity analysis highlights the broader computational benefits of topological fingerprints. Many indices can be computed in linear or near-linear time on sparse graphs, and recent advances in quantum algorithms promise exponential speedups for the \emph{Estrada index}. These properties make topological index kernels particularly attractive in large-scale or resource-constrained environments.


In summary, the proposed extensions to topological index kernels achieve an effective balance between scalability, accuracy, and interpretability. This framework offers viable and versatile alternative to graph kernels, and opens new avenues for efficient, interpretable learning on graph-structured data.



\bibliographystyle{IEEEtran}
\bibliography{article}

\end{document}